\documentclass[conference]{IEEEtran}
\IEEEoverridecommandlockouts
\usepackage{cite}
\usepackage{amsmath,amssymb,amsfonts}
\usepackage{enumitem}
\usepackage{algorithm}
\usepackage{algorithmic}
\usepackage{textcomp}
\usepackage{multirow}
\usepackage{multicol}
\usepackage{lipsum}
\usepackage{float}
\usepackage{dblfloatfix}
\usepackage{mathtools}
\usepackage{xcolor}
\usepackage{array}
\usepackage{url}

\ifCLASSOPTIONcompsoc
\usepackage[caption=false,font=normalsize,labelfont=sf,textfont=sf]{subfig}
\else
\usepackage[caption=false,font=footnotesize]{subfig}
\fi

\def\BibTeX{{\rm B\kern-.05em{\sc i\kern-.025em b}\kern-.08em
    T\kern-.1667em\lower.7ex\hbox{E}\kern-.125emX}}
\begin{document}

\title{An improved online learning algorithm for general fuzzy min-max neural network}

\author{\IEEEauthorblockN{Thanh Tung Khuat}
\IEEEauthorblockA{\textit{Advanced Analytics Institute} \\
\textit{University of Technology Sydney}\\
Sydney, Australia \\
thanhtung.khuat@student.uts.edu.au}
\and
\IEEEauthorblockN{Fang Chen}
\IEEEauthorblockA{\textit{Data Science Centre} \\
\textit{University of Technology Sydney}\\
Sydney, Australia \\
Fang.Chen@uts.edu.au}
\and
\IEEEauthorblockN{Bogdan Gabrys}
\IEEEauthorblockA{\textit{Advanced Analytics Institute} \\
\textit{University of Technology Sydney}\\
Sydney, Australia \\
Bogdan.Gabrys@uts.edu.au}
}

\maketitle

\begin{abstract}
This paper proposes an improved version of the current online learning algorithm for a general fuzzy min-max neural network (GFMM) to tackle existing issues concerning expansion and contraction steps as well as the way of dealing with unseen data located on decision boundaries. These drawbacks lower its classification performance, so an improved algorithm is proposed in this study to address the above limitations. The proposed approach does not use the contraction process for overlapping hyperboxes, which is more likely to increase the error rate as shown in the literature. The empirical results indicated the improvement in the classification accuracy and stability of the proposed method compared to the original version and other fuzzy min-max classifiers. In order to reduce the sensitivity to the training samples presentation order of this new on-line learning algorithm, a simple ensemble method is also proposed.
\end{abstract}


\section{Introduction}
Artificial neural networks (ANNs) are one of the most widely used methods for dealing with classification problems as well as real-world applications \cite{Abiodun18}. However, the main disadvantage of the original ANNs is that they do not have the capability of giving explanations of their predictive results to humans explicitly. This drawback restricts the widespread use of the ANNs for critical domains such as health-care and criminal justice \cite{Rudin19}. In a recent study, Rudin \cite{Rudin19} has highlighted that there is a high demand for interpretable models to substitute black-box models in assisting decision-makers in areas with the requirement of high safety and trust. Hyperbox fuzzy sets can be used to build such interpretable models. Another advantage of adopting hyperbox representations for classifiers is that the model can be trained incrementally. This is an attractive characteristic of hyperbox-based classifiers because it does not require retraining the models periodically. Hence, they can be deployed for applications working in a dynamically changing environment.

\begin{figure}[htbp]
\centerline{\includegraphics[width=0.45\linewidth]{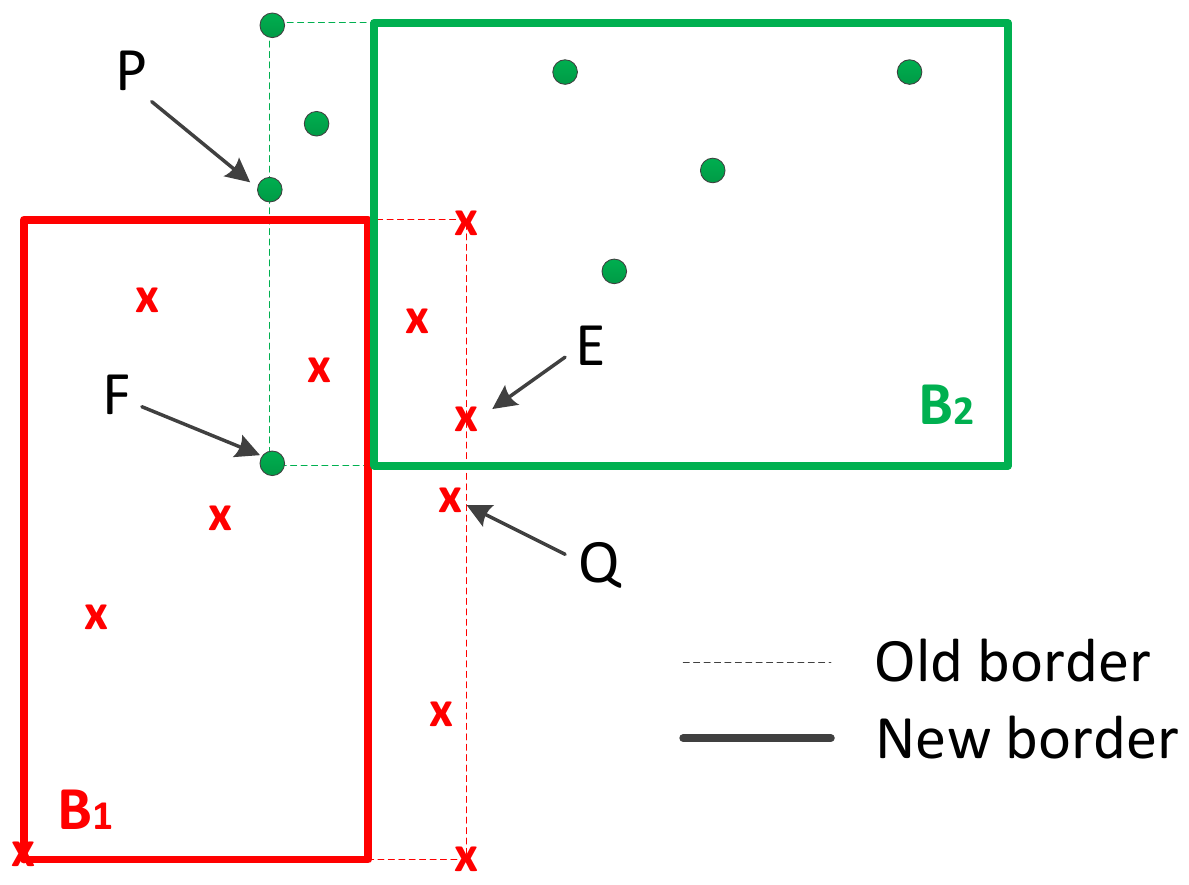}}
\caption{A drawback of the contraction procedure.}
\label{contraction}
\end{figure}

Simpson has proposed to use hyperbox fuzzy sets to form fuzzy min-max neural networks with ideas inspired by the ART neural networks \cite{Carpenter91} to address classification \cite{Simpson92} and clustering \cite{Simpson93} problems. Since then, many studies focused on improving the drawbacks of this class of machine learning algorithms. These enhancements can be divided into two directions, i.e., with and without using the contraction step \cite{Khuat19} to resolve the overlapping regions among hyperboxes. General fuzzy min-max neural network developed by Gabrys and Bargiela \cite{Gabrys00} is a typical representative of improved versions using the contraction procedure. It provides a generalization of fuzzy min-max neural networks with the ability to integrate supervised, semi-supervised, and unsupervised learning in a single framework. However, the use of a contraction process affects the accuracy performance of the GFMM neural network, as shown in \cite{Khuat19} and \cite{Bargiela04} as well as being illustrated in Fig. \ref{contraction}.

As we can see from Fig. \ref{contraction}, some data points such as \textbf{\textit{E}} and \textbf{\textit{F}} belong to wrong hyperboxes after performing the contraction procedure. Other points such as \textbf{\textit{P}} and \textbf{\textit{Q}} are moved to outside of the hyperboxes, and they are more likely to be misclassified. For example, sample \textbf{\textit{P}} covered by hyperbox $\mathbf{B_2}$ before conducting the contraction procedure is now near to hyperbox $\mathbf{B_1}$. Therefore, it will be classified to a red class represented by hyperbox $\mathbf{B_1}$ because the membership degree of \textbf{\textit{P}} in $\mathbf{B_1}$ is higher than that of \textbf{\textit{P}} in $\mathbf{B_2}$.

\begin{figure}[htbp]
\centerline{\includegraphics[width=0.3\linewidth]{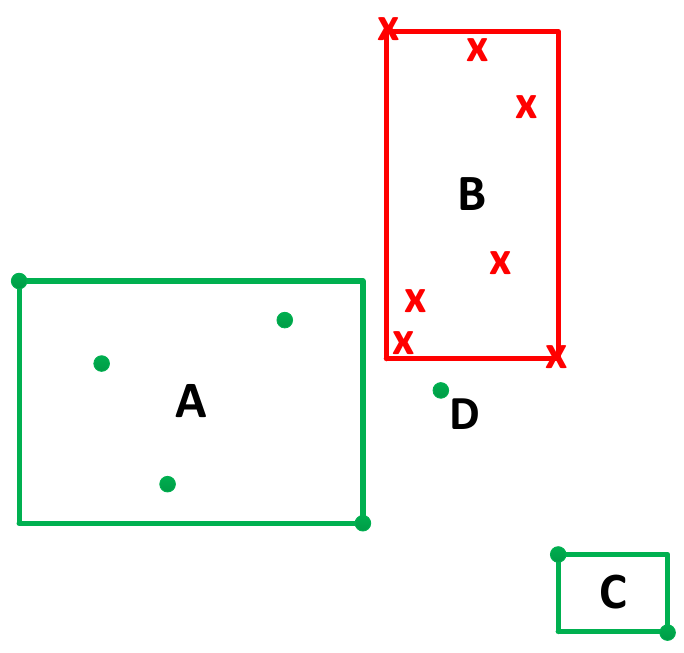}}
\caption{A drawback of the current online learning algorithm for GFMM.}
\label{disadv1}
\end{figure}

Due to the observed undesired effects of the contraction step, other types of fuzzy min-max classifiers such as the inclusion/exclusion fuzzy hyperbox classifier \cite{Bargiela04}, hyperbox classifier with compensatory neurons \cite{Nandedkar07}, data-core-based fuzzy min-max classifier \cite{Zhang11}, and multi-level fuzzy min-max neural network \cite{Davtalab14} did not use the contraction phase to resolve the overlapping regions among hyperboxes. Instead, they deployed a special neuron to handle the overlapping regions. However, these mechanisms make the model architecture complex and increase training time. Therefore, they have been less likely to be applied to large-sized datasets. In this paper, rather than using a special neuron for each overlapping region, we prevent the expansion of hypeboxes if this operation leads to the appearance of overlapping zones.  This principle was very successfully used in the agglomerative learning algorithms in \cite{Gabrys02c}, and it is adopted here in the proposed online learning algorithm. Although the agglomerative learning algorithms are efficient and they do not face the limitations of the overlap resolving step, they use all of the training samples to generate and aggregate hyperboxes repeatedly, so their training time is long. Meanwhile, the online learning algorithm uses a single pass learning mode through the training data, thus its training time is much shorter. In this paper, we propose to use the learning principle of the agglomerative algorithms in a single pass learning mode to construct an improved online learning algorithm for GFMM neural network (IOL-GFMM). A strong point of our proposed method is demonstrated through an example presented in Fig. \ref{disadv1}.

In this figure, the original online learning version of GFMM (Onln-GFMM) will expand hyperbox \textbf{\textit{A}} to cover the new input patter \textbf{\textit{D}}, and then the contraction process is performed to resolve the overlapping region with hyperbox \textbf{\textit{B}}. However, if the IOL-GFMM is used, hyperbox \textbf{\textit{C}} can join the expansion step to cover the input sample \textbf{\textit{D}} because the expansion of hyperbox \textbf{\textit{A}} causes the overlap with \textbf{\textit{B}} representing other class. This process will not lead to any disturbance of the model due to the contraction phase. 

\begin{figure}[!ht]
\centerline{\includegraphics[width=0.32\linewidth]{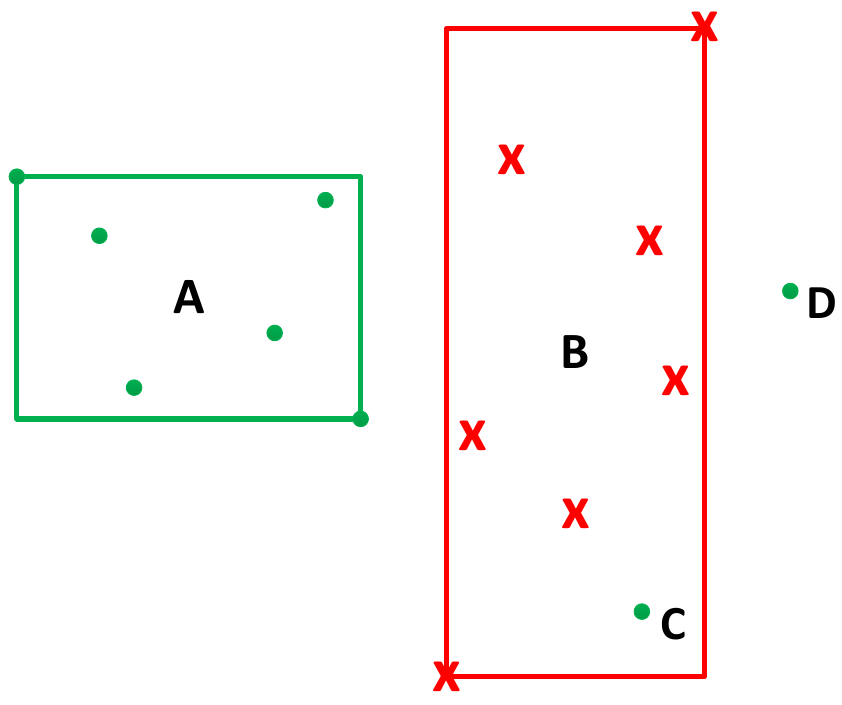}}
\caption{An example of hyperbox-based model and noise.}
\label{noise}
\end{figure}

As mentioned in \cite{Boucheron05}, a classifier is
considered stable if its performance is not varied too much with the small perturbations, e.g. noise, in the training samples. Fuzzy min-max classifiers using online learning algorithms are sensitive to noise, especially when the value of maximum hyperbox size is large. Taking the case shown in Fig. \ref{noise} as an example, with a large value of maximum hyperbox size, hyperbox \textbf{\textit{A}} will be extended to cover the noisy input patterns \textbf{\textit{C}} and \textbf{\textit{D}}. Then, the contraction process conducted can cause the negative disturbance in the learned classifier. In the case of using the IOL-GFMM, hyperbox \textbf{\textit{A}} cannot expand and the patterns \textbf{\textit{C}} and \textbf{\textit{D}} are considered as new hyperboxes without any disturbance. The robustness of the IOL-GFMM to the noisy data will be verified in the experimental part.

Another existing problem of the current Onln-GFMM is that it does not any additional information to allocate the class for unseen data points locating on the decision boundary. In this case, there may exist several winner hyperboxes with the same membership value representing different classes. To cope with this issue, a solution is the use of a Manhattan measure to compute the distance from the input pattern to the central point of each winner hyperbox as introduced in \cite{Upasani19}, and then assigning the input pattern to the winner hyperbox with the smallest distance. However, the central point is the average value of the minimum and maximum coordinates of the hyperbox, so it is also sensitive to noise. In this paper, we use a probability equation generated from cardinality information of each winner hyperbox, as shown in \cite{Gabrys02b}, to decide the suitable class for each input pattern.

With all of the above reasons, we propose the IOL-GFMM. The main distinct point of the proposed method and the original version is that the proposed method does not use the contraction process. The selected hyperbox is only extended to accommodate a new input pattern if this operation does not introduce any overlapping regions with hyperboxes representing other classes. For the classification phase, cardinality information is used to classify unseen data points in the case of existing many winner hyperboxes with the same maximum membership value and belonging to different classes. However, the IOL-GFMM still depends on the order of training data presentation but to a lesser extent than the original Onln-GFMM. With more data seen, the space is becoming more constrained, and new incoming patterns located inside the previously created hyperboxes of other classes cannot be expanded to form hyperboxes. This limitation will be also analyzed in the experimental section and if all training data are stored, we propose a simple ensemble method to tackle this drawback.

The rest of the paper is organized as follows. Section \ref{gfmm} presents the architecture of GFMM neural network and its original online learning algorithm. The improved online learning algorithm to address existing downsides of the current learning approach is mentioned in section \ref{newalg}. Section \ref{exp} describes empirical results and some discussion of the proposed method. Section \ref{conclu} concludes some findings and discusses potential research directions.

\section{General fuzzy min-max neural network} \label{gfmm}
\subsection{An overview architecture of GFMM}
\begin{figure}[!ht]
    \centering
    \includegraphics[width=0.6\linewidth, height=0.3\textwidth]{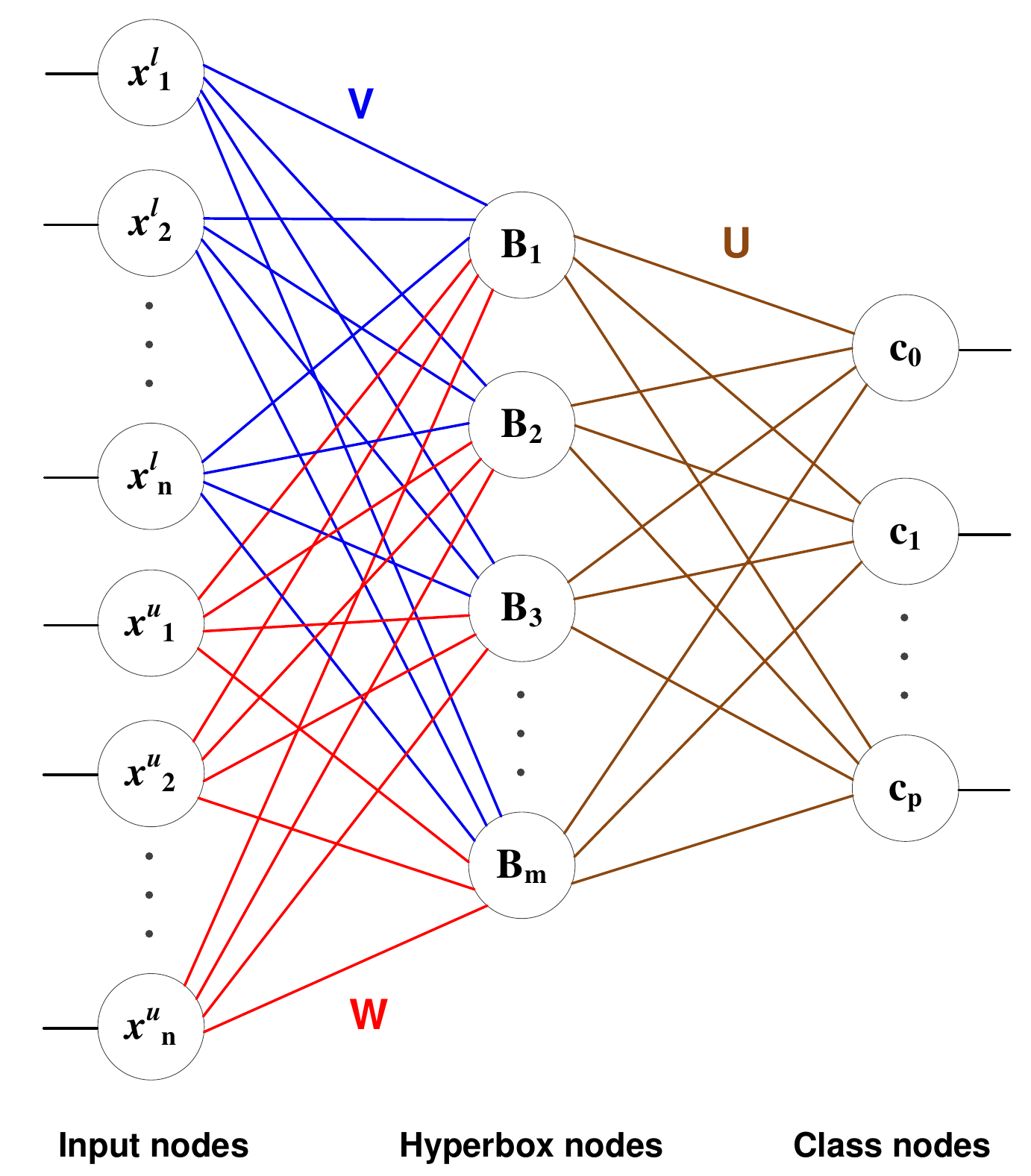}
    \caption{An overview architecture of GFMM.}
    \label{fig_gfmm}
    \vspace{-4mm}
\end{figure}
An overview architecture of GFMM is presented in Fig. \ref{gfmm}. This type of neural network can handle both fuzzy and crisp input data. Therefore, its input layer contains $2n$ nodes, where $n$ is the dimensionality of the input patterns $X = [X^l, X^u]$, specified in an n-dimensional unit cube $\mathbf{I}^n$. The first $n$ nodes in the input layer correspond to $n$ features of the minimum coordinate $X^l$ of the input sample, while the remaining $n$ nodes are $n$ features of the maximum point $X^u$. The hidden layer of the GFMM consists of hyperboxes. Each hyperbox $B_i$ is defined by an ordered set as follows: $B_i = [X, V_i, W_i, b_i(X, V_i, W_i)]$, where $V_i$ and $W_i$ are minimum and maximum points of $B_i$ respectively, $0 \leq b_i \leq 1$ is the membership degree of $B_i$. Minimum and maximum coordinates of all hyperboxes within the hidden layer form two connecting matrices $\mathbf{V}$ and $\mathbf{W}$ between input nodes and hidden nodes. The matrix $\mathbf{V}$ connects the first $n$ input nodes while $\mathbf{W}$ links the remaining $n$ input nodes to the hyperbox nodes. These matrices are adjusted through the learning process of the GFMM. The membership function $b_i$, which is also the activation function of hyperbox nodes, is defined as \eqref{membership}.
\begin{equation}
\small
    b_i(X) = \min\limits_{j = 1}^n(\min([1 - f(x_j^u - w_j, \gamma_j)], [1 - f(v_j - x_j^l, \gamma_j)]))
    \label{membership}
\end{equation}
where $ f(z, \gamma) = \begin{cases} 
1, & \mbox{if } z\gamma > 1 \\
z\gamma, & \mbox{if } 0 \leq z\gamma \leq 1 \\
0, & \mbox{if } z\gamma < 0 \\
\end{cases}
$ is the threshold function and $ \gamma = [ \gamma_1,\ldots, \gamma_n ]$ is a sensitivity parameter showing how fast the membership function decreases.

Hyperbox nodes are fully connected to the output nodes by a binary-valued matrix \textbf{U} of which the elements $u_{ij}$ are calculated using \eqref{outputgfmm}.
\begin{equation}
\label{outputgfmm}
    u_{ij} = \begin{cases}
        1, \quad \mbox{if hyperbox $B_i$ represents class $ c_j $} \\
        0, \quad \mbox{otherwise}
    \end{cases}
\end{equation}
where $ B_i $ is the hyperbox in the hidden layer, and $ c_j $ is the $ j^{th} $ node of the output layer. The transfer function of $ c_j $ is determined by \eqref{eq2}.
\begin{equation}
    \label{eq2}
    c_j = \max \limits_{i = 1}^m {b_i \cdot u_{ij}}
\end{equation}
where $m$ is the number of hyperboxes in the hidden layer. It is noted that node $ c_0 $ is linked to all unlabeled hyperboxes within the hidden layer. The output nodes can produce the fuzzy values computed directly from \eqref{eq2} or crisp values if the node with the highest value is assigned to 1, and the others receive zero values \cite{Gabrys00}.

\subsection{The original online learning algorithm}
The online learning proposed in \cite{Gabrys00} is used to adjust values of two matrices $\mathbf{V}$ and $\mathbf{W}$. This learning algorithm includes three main steps: hyperbox expansion/creation, overlap test, and hyperbox contraction.

For each input sample $X = [X^l, X^u]$ with the class label $l_X$, the algorithm filters all hyperbox candidates with the same label as $l_X$ or unlabelled. Next, the membership degrees between these hyperboxes and $X$ are computed. After that, the hyperbox $B_i$ with the highest membership value is selected, and the condition in \eqref{expcondi} related to maximum hyperbox size ($\theta$) is verified. If the membership value of $B_i$ is one, then $X$ is contained in $B_i$, and nothing is performed.
\begin{equation}
    \label{expcondi}
    \max(w_{ij}, x_j^u) - \min(v_{ij}, x_j^l) \leq \theta, \quad \forall{j \in [1, n]}
\end{equation}
If this condition is satisfied, hyperbox $B_i$ is expanded to cover the input pattern using \eqref{expand} and \eqref{expand2}. If hyperbox $B_i$ is unlabelled, then set $class(B_i) = l_X$.
\begin{align}
    \label{expand}
    v_{ij}^{new} &= \min(v_{ij}^{old}, x_j^l) \\
     \label{expand2}
    w_{ij}^{new} &= \max(w_{ij}^{old}, x_j^u), \quad \forall{j \in [1, n]}
\end{align}
If hyperbox $B_i$ has not met the expansion condition \eqref{expcondi}, the next hyperbox candidate with the second largest membership value is chosen to check condition \eqref{expcondi}. This process is repeated until the selected hyperbox is extended to accommodate the input sample. If no hyperbox candidate satisfies the condition, a new hyperbox is created with min-max coordinates and label identical to the input pattern.

If the expansion procedure at the previous step is performed, the expanded hyperbox $B_i$ is tested the overlap with other hyperboxes $ B_k $ as follows. If $ B_i $ is an unlabelled hyperbox, then it must be checked for overlap with all existing hyperboxes. If $B_i$ is labelled, the overlap test operation is only conducted between $ B_i $ and hyperboxes $ B_k $ representing other classes. For each dimension $j$, four following cases are performed (initially $\delta^{old} = 1$):

\begin{itemize}
    \item $v_{ij} < v_{kj} < w_{ij} < w_{kj}: \delta^{new} = \min(w_{ij} - v_{kj}, \delta^{old}) $
    \item $ v_{kj} < v_{ij} < w_{kj} < w_{ij}: \delta^{new} = \min(w_{kj} - v_{ij}, \delta^{old}) $
    \item $ v_{ij} < v_{kj} \leq w_{kj} < w_{ij}: \\ \delta^{new} = \min(\min(w_{kj} - v_{ij}, w_{ij} - v_{kj}), \delta^{old}) $
    \item $ v_{kj} < v_{ij} \leq w_{ij} < w_{kj}: \\ \delta^{new} = \min(\min(w_{ij} - v_{kj}, w_{kj} - v_{ij}), \delta^{old}) $
\end{itemize}

If $ \delta^{new} < \delta^{old} $, then assign $ \Delta = i$ and $ \delta^{old} = \delta^{new} $ to indicate an overlapping region on the $ \Delta{th} $ dimension, and the testing process is iterated for the next dimension. In contrast, no overlap exists between two hyperboxes, and the contraction process will not been conducted ($\Delta = -1$). If $\Delta \ne -1$, the contraction process is executed on the $\Delta{th}$ dimension to eliminate the overalap region between two hyperboxes. Four cases of this step were presented in \cite{Gabrys00}.

\section{Proposed method} \label{newalg}
To address the identified drawbacks in the original online learning algorithm of the GFMM model mentioned in the introduction section, i.e., the problems of overlap resolving and equal membership degree, we propose an improved version of the original training algorithm. The enhanced algorithm eliminates the contraction process during the training phase, simultaneously using the cardinality information of each hyperbox to support the classification phase. 

\subsection{Training phase}
Similarly to the agglomerative learning algorithm in \cite{Gabrys02c}, the improved learning algorithm of GFMM does not allow the overlap to occur between the expanded hyperbox and any hyperboxes belonging to other classes, so it does not need to use the contraction procedure. The learning process contains two main steps, i.e., expansion/creation of hyperboxes and overlap checking. The details of the proposed method are shown in Algorithm \ref{alg1}. Two main procedures of the algorithm for each input pattern are described as follows:

\paragraph{\textit{Expansion of hyperboxes}} For each input pattern $ X $, we find all existing hyperboxes with the same class as $ X $ or being unlabeled. After that, we compute the membership values of the input pattern to all of these hyperboxes (lines 7-8). Then, we sort all selected hyperboxes in descending order of the membership degrees (line 9). If there is any hyperbox of which the membership value is one, the expansion procedure is not carried out (lines 13-16). Otherwise, we traverse in turns each hyperbox candidate and verify the expansion conditions. If all expansion conditions are met, we update the size of selected hyperbox and the number of samples contained in that hyperbox, and then the expansion step continues with next input patterns (lines 17-27). If all hyperbox candidates are not satisfied with the conditions, we create a new hyperbox to cover the input pattern and add this hyperbox to the current list of hyperboxes (lines 29-31). Two conditions need to be verified, i.e., maximum hyperbox size in \eqref{expcondi} and overlap. If the maximum hyperbox size requirement is met, we then check the non-overlapping condition as follows:

\paragraph{\textit{Overlap test}} The overlap checking occurs between the newly expanded hyperbox and the remaining hyperboxes representing different classes. After expanding the selected hyperbox, if it overlaps with any hyperboxes of other classes, the next hyperbox candidate will be considered. If the extended hyperbox does not overlap with any hyperboxes belonging to other classes, the selected hyperbox will be updated with new size, and the learning process continues with another input patterns. The conditions for the overlap test are the same as in the original GFMM.

Note that we prevent the overlapping areas happening during the hyperbox extension process, but if the input pattern is in the form of hyperbox, the overlap between it and existing hyperboxes representing other classes can still occur. It is due to the fact that we add it directly to the current list of hyperboxes without verifying the overlap. However, such hyperboxes cannot be expanded to cover other patterns because they do not meet the non-overlapping condition. As a result, these hyperboxes are more likely to be removed if we use a pruning step. If not, the classification step using additional cardinality information can still classify the unseen samples correctly because these hyperboxes contain only one sample, which leads to the probability to cover the unseen patterns very small. 

\begin{algorithm} [!ht]
	\caption{The improved learning algorithm of GFMM} \label{alg1}
	\footnotesize{
	\begin{algorithmic} [1]
	    \REQUIRE
	    \item[]
	    \begin{itemize}
	        \item $ \theta $: The maximum hyperbox size
	        \item $ \gamma $: The speed of decreasing of the membership function
	    \end{itemize}
	    \ENSURE
	    \item[]
	    A list $ \mathcal{H} $ of hyperbox fuzzy sets containing minimum-maximum values and classes
	    \item[]
	    \STATE Initialize an empty list of hyperboxes: min-max values $ \mathcal{V} = \mathcal{W} = \varnothing $, hyperbox classes: $ \mathcal{L} = \varnothing $
	    \FOR{each input pattern $X = [X^l, X^u, l_X]$}
    	    \STATE $ n \leftarrow $ The number of dimensions of $ X $
    	    \IF{$ \mathcal{V} = \varnothing$}
    	        \STATE $ \mathcal{V} \leftarrow X^l; \quad \mathcal{W} \leftarrow X^u; \quad \mathcal{L} \leftarrow l_X $
    	    \ELSE
    	        \STATE $\mathcal{H}_1 = [\mathcal{V}_1, \mathcal{W}_1, \mathcal{L}_1] \leftarrow $ Find hyperboxes in $ \mathcal{H} = [\mathcal{V}, \mathcal{W}, \mathcal{L}] $ representing the same class as $ X $ or being unlabeled
        	    \STATE $ \mathcal{M} \leftarrow $ \textbf{ComputeMembershipValue}($X, \mathcal{V}_1, \mathcal{W}_1, \mathcal{L}_1$)
        	    \STATE $ \mathcal{H}_d \leftarrow $ \textbf{SortByDescending}($ \mathcal{H}_1, \mathcal{M}(\mathcal{H}_1)$)
        	    \STATE Set $ \overline{\mathcal{H}_1} \leftarrow \mathcal{H} \setminus \mathcal{H}_1 $
        	    \STATE $ flag \leftarrow false $
    	        \FOR{each $h = [V_h, W_h, l_h] \in \mathcal{H}_d$}
    	            \IF{$\mathcal{M}(h)$ = 1}
    	                \STATE $flag = true$
    	                \STATE \textbf{break}
    	            \ENDIF
    	            \IF{$\max(w_{hj}, x_{j}^u) - \min(v_{hj}, x_{j}^l) \leq \theta, \forall {j \in [1, n]} $}
    	            \STATE $ W_h^t \leftarrow \max(W_h, X^u); \quad V_h^t \leftarrow \min(V_h, X^l)$
    	            \STATE $isOver \leftarrow $ \textbf{IsOverlap}($W_h^t, V_h^t, \overline{H_1}$)
    	            \IF{$ isOver = false$}
    	                \STATE $ V_h \leftarrow V_h^t; \quad W_h \leftarrow W_h^t$
    	                \STATE $ l_h \leftarrow l_X $ if $ l_h $ is unlabeled and $ l_X $ is labeled
    	                \STATE $ flag \leftarrow true$
    	                \STATE Increase the number of samples contained in $h$
    	                \STATE \textbf{break}
    	            \ENDIF
    	            \ENDIF
    	        \ENDFOR
    	        \IF{$flag = false$}
    	            \STATE $ \mathcal{V} \leftarrow \mathcal{V} \cup X^l; \quad \mathcal{W} \leftarrow \mathcal{W} \cup X^u; \quad \mathcal{L} \leftarrow L \cup l_X $
    	        \ENDIF
    	    \ENDIF
    	\ENDFOR
		\RETURN $ \mathcal{H} = [\mathcal{V}, \mathcal{W}, \mathcal{L}] $
	\end{algorithmic}
	}
\end{algorithm}

\subsection{Classification phase}
For an unseen input pattern $X$, the membership values between $X$ and hyperboxes of the trained model are computed. Then, the input $X$ is classified to the hypebox with the highest membership value. If many hyperboxes representing $ K $ different classes have the same maximum membership value ($ b_{win}$). If $ b_{win} = 1$ and $\exists i: n_i = 1$, then the prediction class of $X$ is the class of $B_i$. Otherwise, we will use an additional formula to find the right class for the input pattern as shown in \eqref{probcard} \cite{Gabrys02b}. The final class of an input pattern is the class $c_k$ with the highest value of $\mathcal{P}(c_k|X) $.

\begin{equation}
    \label{probcard}
    \mathcal{P}(c_k|X) = \cfrac{\sum_{j \in \mathcal{I}_{win}^k} n_j \cdot b_j}{\sum_{i \in \mathcal{I}_{win}} n_i \cdot b_i }
\end{equation}
where $k \in [1, K]$ and $ \mathcal{I}_{win} = \{ i, \mbox{if } b_i = b_{win} \}$ is a set of indexes of all hyperbox with the same highest membership value, $ I_{win}^k = \{ j, \mbox{if } class(B_j) = c_k \mbox{ and } b_i = b_{win} \}$ is a subset of $ I_{win} $ with indexes for the $k^{th}$ class, and $n_i$ is the number of samples contained in the hyperbox $ B_i$. 

\section{Experiments} \label{exp}
Experiments in this part were conducted on popular datasets stored in UCI repository \cite{Dua19}. Several descriptions of the used datasets are shown in Table \ref{data}. Because these datasets have a small size, we adopted the density-preserving sampling (DPS) method \cite{Budka13} to split the datasets aiming to preserve the data density and class shapes for folds employed in training and testing phases. Each dataset was partitioned into four folds using the DPS approach. Three folds were used as training data to build the learning model, and the remaining fold was deployed to evaluate the performance of the trained model. This process was repeated until all folds were used as testing data. The average testing error rates on four folds are reported in this paper.

\begin{table}[!ht]
\centering
\caption{Descriptions of Datasets} \label{data}
\scriptsize
\begin{tabular}{|c|l|c|c|c|}
\hline
\textbf{ID} & \textbf{Dataset}      & \multicolumn{1}{l|}{\textbf{\# samples}} & \multicolumn{1}{l|}{\textbf{\# features}} & \multicolumn{1}{l|}{\textbf{\# classes}} \\ \hline
1           & Blood transfusion     & 748                                      & 4                                         & 2                                        \\ \hline
2           & Breast Cancer Coimbra & 116                                      & 9                                         & 2                                        \\ \hline
3           & Haberman              & 306                                      & 3                                         & 2                                        \\ \hline
4           & Heart                 & 270                                      & 13                                        & 2                                        \\ \hline
5           & Page blocks           & 5473                                     & 10                                        & 5                                        \\ \hline
6           & Landsat Satellite     & 6435                                     & 36                                        & 6                                        \\ \hline
7           & Waveform (v1)         & 5000                                     & 21                                        & 3                                        \\ \hline
8           & Yeast                 & 1484                                     & 8                                         & 10                                       \\ \hline
9           & Spherical\_5\_2           & 250                                      & 2                                         & 5 \\ \hline
\end{tabular}
\end{table}

\subsection{Comparison of the performance and complexity of the proposed and original learning algorithms of GFMM}\label{cmpper_ori}
This experiment is to assess the performance and complexity between the proposed method and existing learning algorithms of GFMM including online learning (Onln-GFMM) and accelerated agglomerative (batch) learning (AGGLO-2). In addition to the membership value, the original online learning algorithm does not use any additional information to support the classification step. Therefore, in the case that there are many hyperboxes with the same maximum membership degree, we will select randomly among classes of the winner hyperboxes to return as the predicted class. To strengthen the comparison with the proposed method, in the original online learning algorithm, we use the hyperbox central point (the average value of maximum and minimum coordinates) and a Manhattan distance measure (Onln-GFMM + Manhattan) to handle the case of many winner hyperboxes as shown in \cite{Upasani19}. The Manhattan distance from the input pattern to central points of the winner hyperboxes are calculated. Then, the input pattern will be assigned the hyperbox with the minimum value of the Manhattan distance. For AGGLO-2, we use \eqref{probcard} to find the predictive class in the case of many winner hyperboxes.

\begin{figure*}
	\centering
	\subfloat[Error rate on the \textit{Haberman} dataset\label{sub_Haberman_err}]{%
		\includegraphics[width=0.32\linewidth, height   =0.25\linewidth]{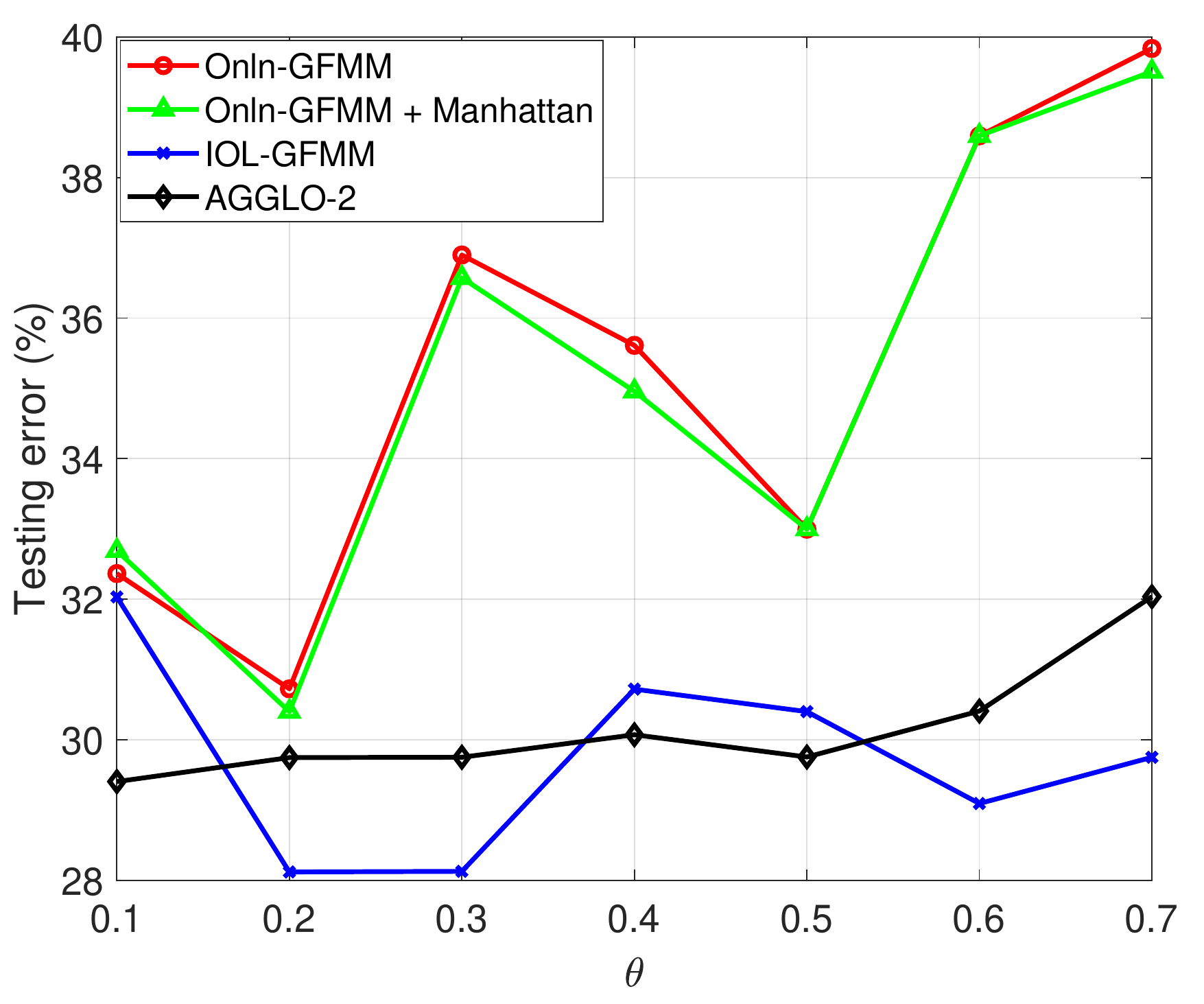}
	}
	\hfill
	\subfloat[Model complexity on the \textit{Haberman} dataset \label{sub_Haberman_complx}]{%
		\includegraphics[width=0.32\linewidth, height   =0.25\linewidth]{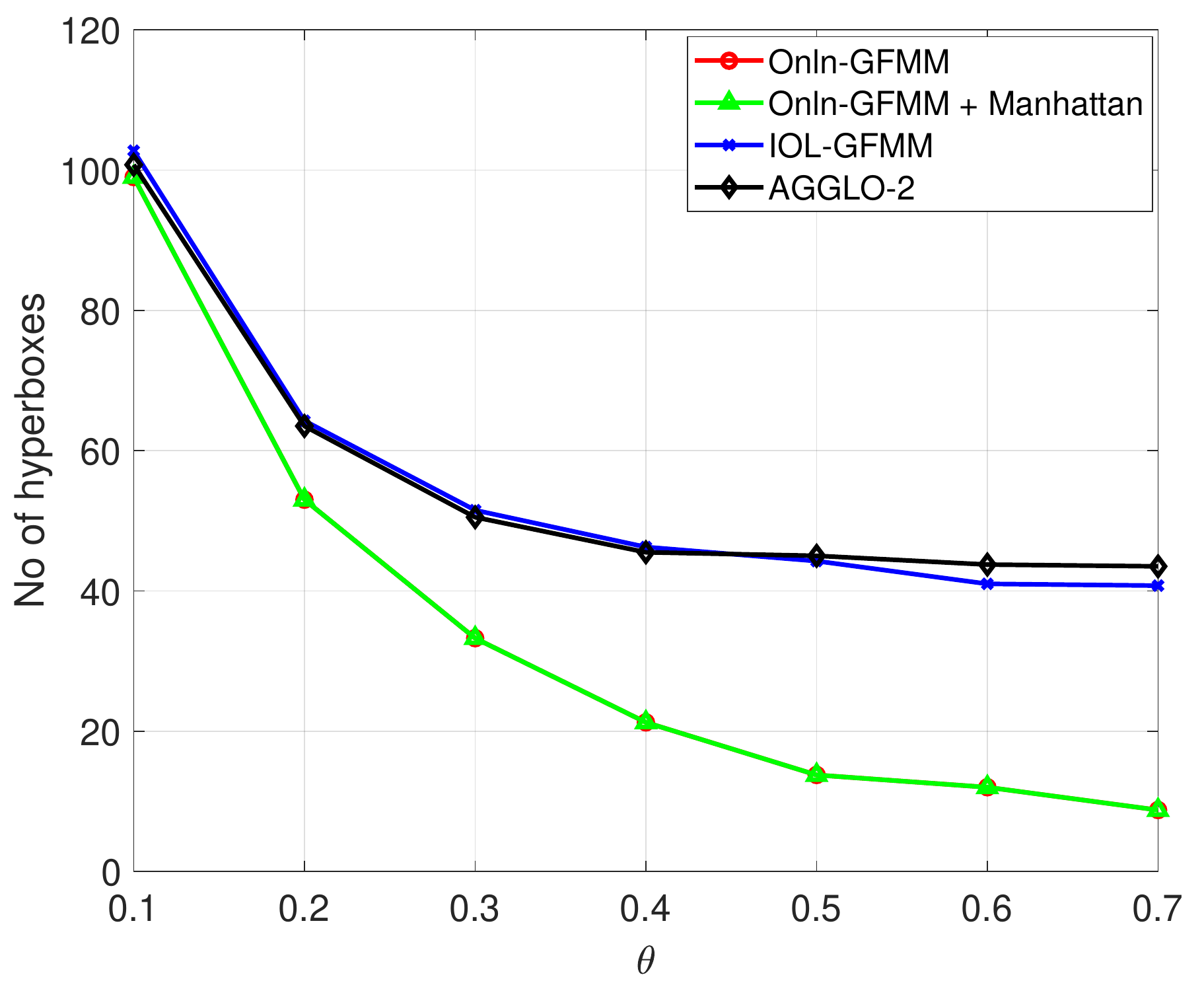}
	}
	\hfill
	\subfloat[Error rate on the \textit{Page blocks} dataset\label{sub_Pageblock_err}]{%
		\includegraphics[width=0.32\linewidth, height   =0.25\linewidth]{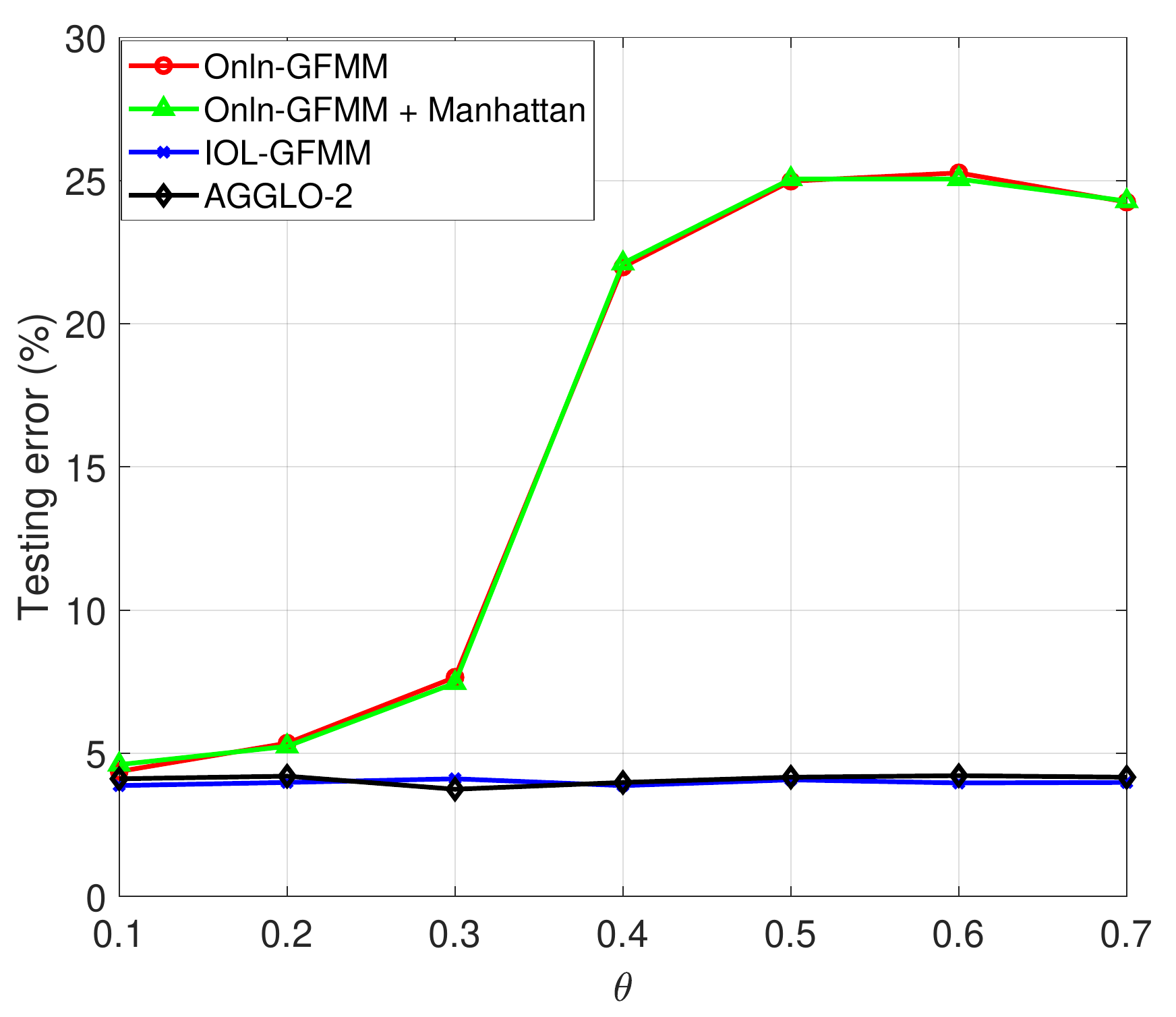}
	}
	\hfill
	\subfloat[Model complexity on the \textit{Page blocks} dataset\label{sub_Pageblock_complx}]{%
		\includegraphics[width=0.32\linewidth, height   =0.25\linewidth]{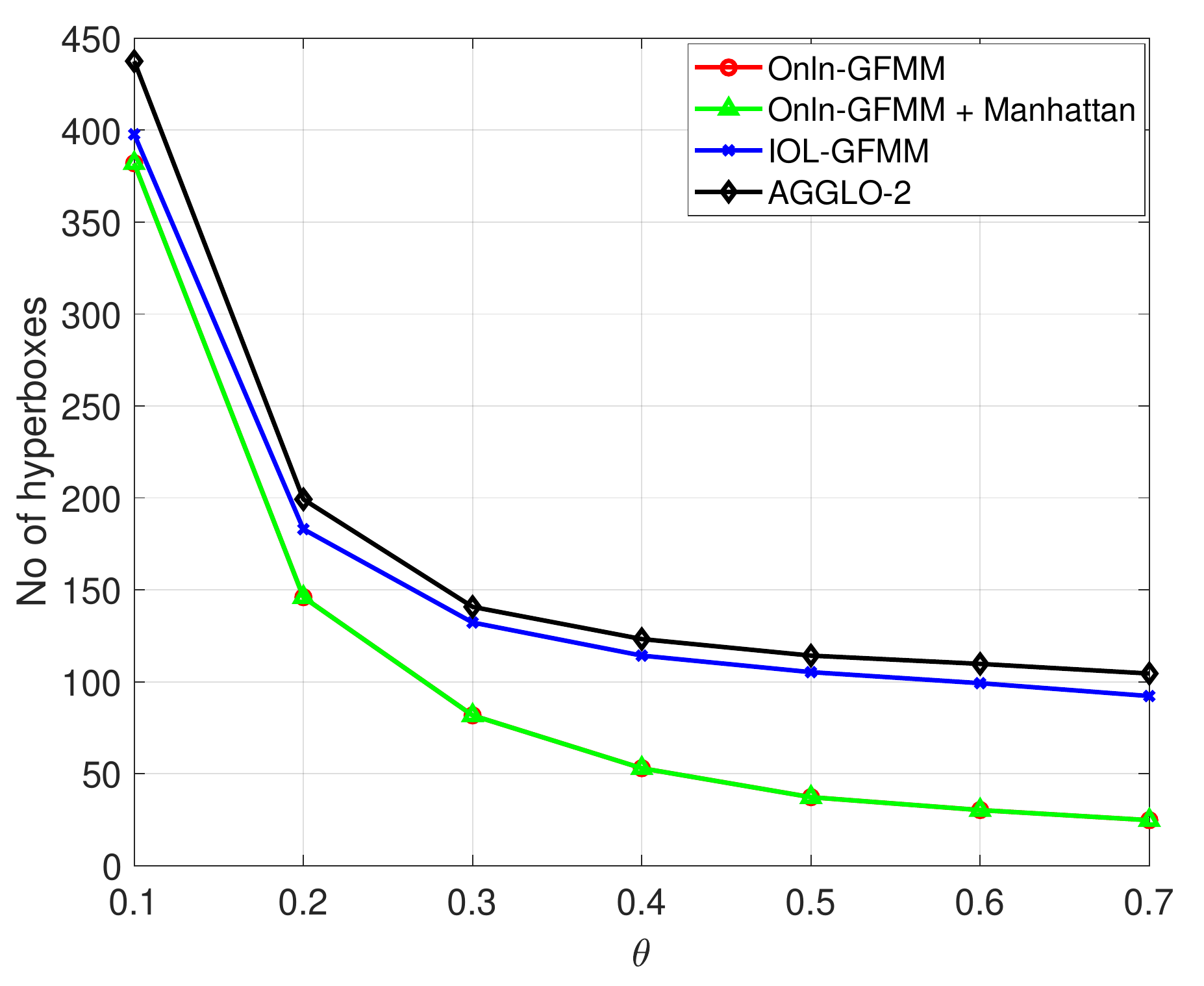}
	}
	\hfill
	\subfloat[Error rate on the \textit{Landsat Satellite} dataset\label{sub_Lansat_err}]{
		\includegraphics[width=0.32\linewidth, height   =0.25\linewidth]{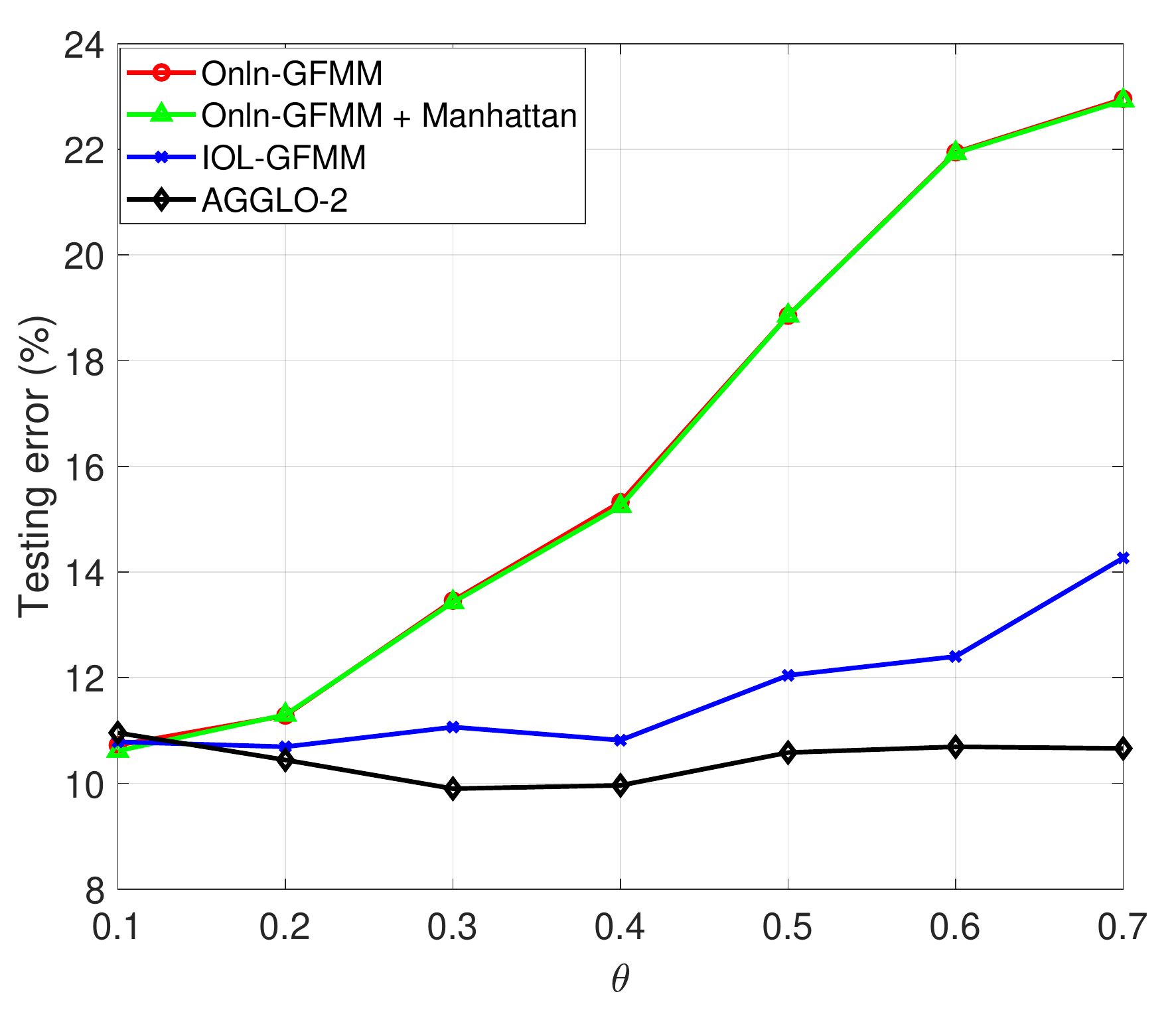}
	}
	\hfill
	\subfloat[Model complexity on the \textit{Landsat Satellite} datase\label{sub_Lansat_complx}]{
		\includegraphics[width=0.32\linewidth, height   =0.25\linewidth]{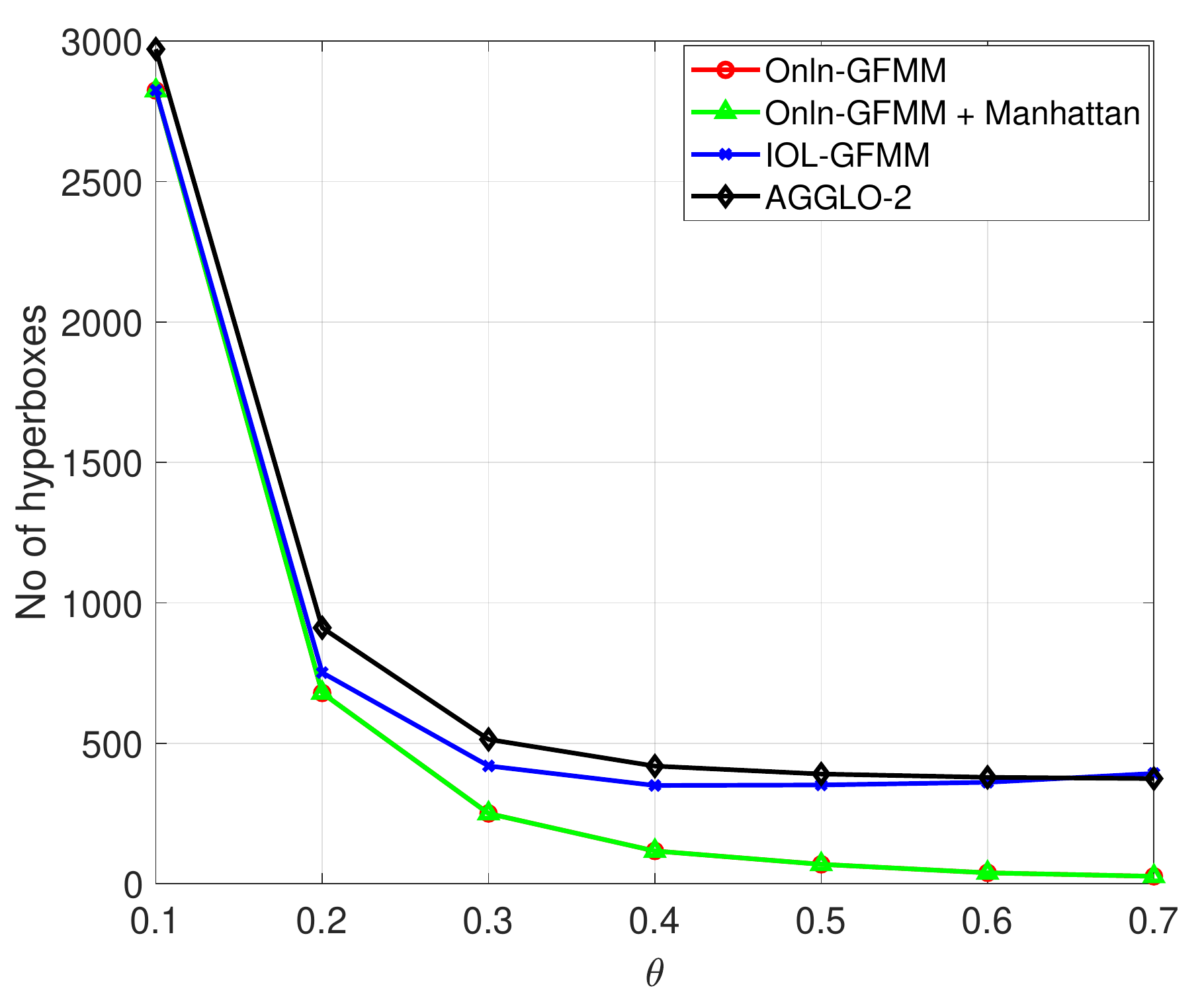}
	}
	\caption{The performance and model complexity of the proposed and original online learning algorithm.}
	\label{compare_per_complx}
\end{figure*}

This experiment aims to assess the influences of the maximum hyperbox size parameter ($\theta$) on the classification accuracy, model complexity, and training time of the proposed and original learning algorithms using different datasets. Three representative datasets, i.e., \textit{Haberman}, \textit{Page blocks}, and \textit{Landsat Satellite}, representing the diversity in the numbers of samples, features, and classes, were selected for this purpose. The values of $\theta$ were increased from 0.1 to 0.7, and at each threshold, the models were trained and assessed repeatedly four times on four different training and testing DPS folds. Fig. \ref{compare_per_complx} shows the average test error rates and complexity (the number of generated hyperboxes) of models on three datasets at different values of $\theta$.

It can be observed that the GFMM classifier trained using the proposed online learning algorithm produces smaller average testing errors compared to the models using the original learning algorithms, especially in the case of employing high values of $\theta$. The use of the Manhattan distance measure contributes to reducing the testing error rates of the model using the old online learning algorithm. While the performances of GFMM trained by the old online learning algorithm are severely affected by the increase of maximum hyperbox sizes, the proposed online learning algorithm only causes a slight increase in the testing error rates of model when using large values of $\theta$. These facts can be explained by observing the number of generated hyperboxes at each threshold of $\theta$. At small values of $\theta$, the complexity of the model using the proposed algorithm is nearly equivalent to that of the classifier using the original approach. At higher settings of $\theta$, the numbers of generated hyperboxes of three methods are reduced, but the complexity of the model trained by the improved algorithm is much higher than the models using the original method. As a result, knowledge coded in the classifier using the improved method is still maintained, so the test accuracy rates are not reduced considerably. In other words, the proposed learning algorithm is efficient in lowering the misclassification errors when increasing the maximum hyperbox size. Compared to batch learning algorithm AGGLO-2, error rates of IOL-GFMM are only slightly higher in some thresholds, while its complexity is usually smaller.

\begin{table*}[!ht]
\centering
\caption{Average Training Time (in Seconds) of GFMM Classifiers Using the Original and Improved Learning Algorithms \break (Smaller Values are Better and Highlighted in Bold)} \label{trainingtime}
\scriptsize
\begin{tabular}{|c|l|l|c|c|c|c|c|c|c|}
\hline
\textbf{ID}                 & \textbf{Dataset}                                                                      & \textbf{Model}      & $\theta = 0.1$ & $\theta = 0.2$ & $\theta = 0.3$ & $\theta = 0.4$ & $\theta = 0.5$ & $\theta = 0.6$ & $\theta = 0.7$ \\ \hline
\multirow{3}{*}{1} & \multirow{3}{*}{Page blocks}                                                & Onln-GFMM     & 3.8717         & 1.9758         & 1.3024         & 1.0879         & 1.1667         & 1.0385         & 1.0290          \\ \cline{3-10} 
                   &                                                                              & IOL-GFMM & \textbf{0.8238}         & \textbf{0.4376}         & \textbf{0.3484}         & \textbf{0.3432}         & \textbf{0.3223}         & \textbf{0.3306}         & \textbf{0.3313}         \\
\cline{3-10}
& & AGGLO-2 & 15.6012 & 6.3211 & 5.1393 & 5.2248 & 5.8189 & 6.2523 & 6.0002 \\ \hline
\multirow{3}{*}{2} & \multirow{3}{*}{Landsat Satellite} & Onln-GFMM     & 67.1135        & 27.5141       & 7.4381         & 3.3578         & \textbf{2.2326}         & \textbf{1.5856}         & \textbf{1.1288}         \\ \cline{3-10} 
                   &                                                                              & IOL-GFMM & \textbf{8.1266}         & \textbf{2.0505}         & \textbf{1.7381}         & \textbf{2.1205}         & 2.4473        & 2.7987        & 3.2385         \\
\cline{3-10}
& & AGGLO-2 & 867.9369 & 68.8635 & 46.1044 & 54.4781 & 58.4633 & 66.5776 & 60.4559 \\
                   \hline
\multirow{3}{*}{3} & \multirow{3}{*}{Haberman}                                                    & Onln-GFMM     & 0.0794         & 0.0649         & 0.0547         & 0.0343         & \textbf{0.0272}         & 0.0371         & \textbf{0.0248}         \\ \cline{3-10} 
                   &                                                                              & IOL-GFMM & \textbf{0.0376}         & \textbf{0.0290}         & \textbf{0.0275}         & \textbf{0.0339}         & 0.0313         & \textbf{0.0309}         & 0.0346 \\
\cline{3-10}
& & AGGLO-2 & 0.2493 & 0.1604 & 0.1219 & 0.1385 & 0.1571 & 0.1869 & 0.1812 \\ \hline
\end{tabular}
\end{table*}

Table \ref{trainingtime} shows the average training time of the original and proposed learning algorithms on three datasets. In general, the training time of the IOL-GFMM is much faster than the original online learning algorithm when the maximum hyperbox size is small (usually $\leq 0.4$). It is due to the fact that the proposed learning algorithm reduces the number of expandable hypebox candidates as well as without using the contraction process. In the case that the value of $\theta$ is high and the number of data points in the training set is large such as \textit{Landsat Satelite} dataset, the training time of the proposed method is longer than that of the original learning algorithm. This fact is not surprising since the number of generated hyperboxes at high values of $\theta$ using the proposed method is much larger than that using the original learning approach (see Fig. \ref{compare_per_complx}-(f)). Hence, the overlap test procedure in the proposed method has to run on much more hyperboxes, so it increases the training time. Furthermore, the training time of online learning algorithms is much faster than that of the agglomerative learning algorithm. These figures show that the IOL-GFMM algorithm is competitive to the batch learning algorithm while its training time is much faster.

\subsection{Evaluation the robustness of the proposed method to noise}
This experiment is to prove the robustness, as shown in the example in Fig. \ref{noise}, of the improved online learning algorithm for GFMM. To evaluate the effectiveness of the IOL-GFMM and the pruning operation in dealing with noise in the learning process with single pass through, three representative datasets, i.e., \textit{Habermann}, \textit{Landsat Satellite}, and \textit{Page blocks} as shown in the subsection \ref{cmpper_ori}, were used. Each dataset was split into four folds using the DPS method; two folds were used for building the models, one fold for pruning, and one fold for testing. The training and validation sets were corrupted with 5\%, 10\%, and 15\% of the total number of samples changed their class labels randomly. The experiments were repeated four times for four different testing folds to obtain the testing errors before and after conducting the pruning step. This paper uses a simple pruning operation, in which hyperboxes with predictive accuracy lower than 0.5 are eliminated \cite{Khuat19b}.

We conducted experiments using a small value of $\theta = 0.1$ and a large value of $\theta = 0.7$. Table \ref{noise01} presents the obtained results with $\theta = 0.1$, and Table \ref{noise07} is testing error results with $\theta = 0.7$.

\begin{table}[!ht]
\centering
\caption{Testing Errors on Noisy Datasets in the Case of Using $\theta = 0.1$} \label{noise01}
{\scriptsize
\begin{tabular}{|c|l|l|l|l|l|l|}
\hline
\multirow{2}{*}{\textbf{\%Noise}} & \multicolumn{2}{c|}{\textbf{GFMM}}                                        & \multicolumn{2}{c|}{\textbf{GFMM-Manhattan}}                              & \multicolumn{2}{c|}{\textbf{IOL-GFMM}}                                    \\ \cline{2-7} 
                                  & \multicolumn{1}{c|}{\textbf{B. Pr}$^{\mathrm{a}}$} & \multicolumn{1}{c|}{\textbf{A. Pr}$^{\mathrm{b}}$} & \multicolumn{1}{c|}{\textbf{B. Pr}} & \multicolumn{1}{c|}{\textbf{A. Pr}} & \multicolumn{1}{c|}{\textbf{B. Pr}} & \multicolumn{1}{c|}{\textbf{A. Pr}} \\ \hline
\multicolumn{7}{|l|}{\textbf{Haberman}} \\ \hline
0  & 28.460  & 27.474 & 30.421 & 30.712 & 28.132 & 31.707                              \\ \hline
5  & 30.404  & 28.777 & 29.755 & 29.401 & 28.448 & 27.781                              \\ \hline
10 & 32.045  & 32.702 & 32.698 & 31.694 & 31.066 & 31.037                              \\ \hline
15 & 34.330  & 31.374 & 35.646 & 34.3175 & 32.698 & 33.672                              \\ \hline
\multicolumn{7}{|l|}{\textbf{Landsat Satellite}} \\ \hline
0 & 11.313 & 13.458 & 11.236  & 11.064  & 11.360  & 11.313                              \\ \hline
5 & 12.945 & 14.080  & 12.883  & 12.370 & 12.681  & 12.417                              \\ \hline
10 & 15.090 & 15.214  & 15.027 & 14.577 & 14.483 & 14.918                              \\ \hline
15 & 17.250 & 17.125  & 17.219  & 14.266 & 16.503 & 14.499                              \\ \hline
\multicolumn{7}{|l|}{\textbf{Page blocks}}  \\ \hline
0   & 4.404 & 4.824  & 4.422  & 6.760  & 3.874  & 7.326                               \\ \hline
5   & 14.142  & 13.959 & 14.306 & 8.643  & 4.952  & 7.875                               \\ \hline
10  & 20.263  & 19.880 & 20.501  & 11.565  & 6.085  & 8.186                               \\ \hline
15  & 27.608  & 26.768 & 27.828  & 15.183  & 8.240 & 7.528                               \\ \hline
\multicolumn{7}{|l|}{$^{\mathrm{a}}$B. Pr: Before pruning;   $^{\mathrm{b}}$A. Pr: After pruning}                                                                                                                                                                                                 \\ \hline
\end{tabular}
}
\end{table}

It can be seen that the GFMM models using the old online learning algorithms with large-sized hyperboxes ($\theta = 0.7$) are sensitive to noise because the testing error increases significantly at the level of 15\% noise on three considered datasets, while they are less insensitive to noise in the case of small-sized hyperboxes. Meanwhile, the IOL-GFMM is less affected by noise even in the case of employing a large value of $\theta$ as the error rates only increase $< 7\%$ when there are noisy samples in the training data. In general, the IOL-GFMM is more stable than the original GFMM, and this claim is demonstrated via the \textit{Landsat Satellite} and \textit{Page blocks} datasets. We can also see that the pruning procedure contributes to decreasing the error rates of models trained by the original online learning algorithms in some cases. Nevertheless, it is not effective for IOL-GFMM, especially when using a high maximum hyperbox size threshold because it leads to an increase in the testing error. In general, the pruning procedure is efficient on the model with many small-sized hyperboxes. With a small value of $\theta$ and high rate of noise, the pruning step is very useful when it contributes to a significant reduction of error rates. However, in the case of using large values of $\theta$, the pruning step is ineffective for the IOL-GFMM model. This fact confirms that the generated hyperboxes using the proposed method are highly valuable ones, and the removal of these hyperboxes leads to the loss of learned knowledge for unusual patterns.

\begin{table}[!ht]
\centering
\caption{Testing Errors on Noisy Datasets in the Case of Using $\theta = 0.7$} \label{noise07}
{\scriptsize
\begin{tabular}{|c|l|l|l|l|l|l|}
\hline
\multirow{2}{*}{\textbf{\%Noise}} & \multicolumn{2}{c|}{\textbf{GFMM}}                                        & \multicolumn{2}{c|}{\textbf{GFMM-Manhattan}}                              & \multicolumn{2}{c|}{\textbf{IOL-GFMM}}                                    \\ \cline{2-7} 
                                  & \multicolumn{1}{c|}{\textbf{B. Pr}$^{\mathrm{a}}$} & \multicolumn{1}{c|}{\textbf{A. Pr}$^{\mathrm{b}}$} & \multicolumn{1}{c|}{\textbf{B. Pr}} & \multicolumn{1}{c|}{\textbf{A. Pr}} & \multicolumn{1}{c|}{\textbf{B. Pr}} & \multicolumn{1}{c|}{\textbf{A. Pr}} \\ \hline
\multicolumn{7}{|l|}{\textbf{Haberman}} \\ \hline
0  & 28.439  & 28.439 & 28.439  & 39.807 & 28.777 & 44.412                              \\ \hline
5  & 28.435  & 28.435 & 28.435  & 39.807 & 30.071 & 37.239                              \\ \hline
10 & 33.002  & 33.002 & 33.002  & 41.554 & 33.356 & 38.901                              \\ \hline
15 & 47.374  & 47.374 & 47.374  & 37.838 & 34.638 & 37.615                              \\ \hline
\multicolumn{7}{|l|}{\textbf{Landsat Satellite}}  \\ \hline
0  & 24.335  & 24.351 & 24.195 & 35.429  & 14.437 & 21.322                              \\ \hline
5  & 42.269  & 42.284 & 42.269 & 50.210  & 14.918 & 21.259                              \\ \hline
10  & 56.736 & 56.643 & 56.674 & 59.346  & 17.124 & 21.072                              \\ \hline
15  & 59.053 & 58.913 & 58.960  & 66.994 & 20.357 & 25.409                              \\ \hline
\multicolumn{7}{|l|}{\textbf{Page blocks}} \\ \hline
0  & 28.360  & 28.396  & 28.378  & 42.878  & 4.002  & 28.941                                \\ \hline
5  & 64.387  & 64.478  & 64.387  & 9.757  & 4.532  & 17.304                               \\ \hline
10 & 77.928  & 77.947  & 78.001  & 51.993  & 5.664   & 24.795                                \\ \hline
15 & 81.328  & 81.310   & 81.291  & 81.602  & 5.865  & 24.504                                \\ \hline
\multicolumn{7}{|l|}{$^{\mathrm{a}}$B. Pr: Before pruning;   $^{\mathrm{b}}$A. Pr: After pruning} \\ \hline
\end{tabular}
}
\end{table}

\subsection{Comparison of the performance of the proposed method to other fuzzy min-max classifiers}
This experiment is to compare the performance of the GFMM model using the improved online learning algorithm to other types of fuzzy min-max neural networks such as the GFMM using the old online learning (Onln-GFMM) and agglomerative learning (AGGLO-2) \cite{Gabrys02c} algorithms, fuzzy min-max neural network (FMNN) \cite{Simpson92}, enhanced fuzzy min-max neural network (EFMNN) \cite{Mohammed15}, and enhanced fuzzy min-max neural network using K-nearest hyperbox expansion rule (KNEFMNN) \cite{Mohammed17}. To conduct a fair comparison, we used 3-fold DPS cross-validation and a grid-search procedure to find the best value of $\theta$ in the range of $[0.1, 0.15, \ldots, 0.65, 0.7]$ for each classifier. As for KNEFMNN, the value of $K$ was searched in the range of [2, 10]. For each training dataset including three DPS folds, a fold was used as a validation set, and two remaining folds were deployed to train the model. Next, the performance of the trained classifier was assessed on the validation fold. This process was iterated three times for three different validation folds for each set of parameters. The parameters which resulted in the lowest average validation error rate were used to train the final classifier using three DPS folds. Finally, the final model was executed on the remaining DPS testing fold. This process was repeated four times for four different DPS testing folds, and the average testing error is reported in Table \ref{avgper}.

\begin{table*}[!ht]
\centering
\caption{The Lowest Average Testing Errors (\%) of Models on Nine Datasets (Smaller Values are Better and Shown in Bold)} \label{avgper}
\scriptsize
\begin{tabular}{|c|l|r|r|r|r|r|r|r|}
\hline
\textbf{ID} & \textbf{Dataset}    & \textbf{Onln-GFMM} & \textbf{Onln-GFMM + Manhattan} & \textbf{IOL-GFMM} & \textbf{AGGLO-2} & \textbf{FMNN} & \textbf{EFMNN} & \textbf{KNEFMNN} \\ \hline
1           & Blood transfusion  & 34.35825      & 33.28875                & \textbf{23.797}    & 24.59875  & 30.8825        & 33.28875         & 32.353         \\ \hline
2           & Breast Cancer Coimbra & 28.448       & 28.448                & 32.7585 & 26.724           & 30.172        & 37.069         & \textbf{25.862}           \\ \hline
3           & Haberman            & 31.703      & 32.032                & \textbf{28.76775}  & 31.374    & 39.8325        & 34.00975       & 30.3785           \\ \hline
4           & Heart               & 23.32625      & 19.24375                & 22.218 & 22.5965          & \textbf{18.15725}      & 19.99       & 18.8815          \\ \hline
5           & Page blocks        & 4.54975       & 4.6045                 & \textbf{3.7455}      & 3.91      & 6.99775       & 4.915        & 4.44          \\ \hline
6           & Landsat Satellite             & 11.08        & 11.01775                & 10.816    & \textbf{10.30275}        & 17.54475       & 11.82625        & 10.8935           \\ \hline
7           & Waveform            & 18           & 18                      & 18.7      & \textbf{17.92}        & 22.2         & 21.22          & 20.78            \\ \hline
8           & Yeast               & 42.116       & 42.18475                  & 41.85025 & 42.989          & 51.62025      & 40.8355        & \textbf{39.962}         \\ \hline
9 & Spherical\_5\_2 & 1.20325 & 1.20325 & 1.20325 & 1.20325 & \textbf{1.19675} & 2.40675 & 1.60025 \\ \hline

\end{tabular}
\end{table*}

\begin{table} [!ht]
\centering
\caption{The Average Rank of the Performance of Models through Nine Datasets (Best Value Is Highlighted in Bold)} \label{avgrank}
\scriptsize
\begin{tabular}{|c|l|c|}
\hline
\textbf{ID} & \textbf{Model}                   & \textbf{Average rank} \\ \hline
1           & Onln-GFMM                        & 4.50                  \\ \hline
2           & Onln-GFMM + Manhattan & 4.11                  \\ \hline
3           & FMNN                                 & 5.0                  \\ \hline
4           & EFMNN                                & 5.5                    \\ \hline
5           & KNEFMNN                              & 3                     \\ \hline
6           & AGGLO-2            & \textbf{2.94} \\ \hline
7           & IOL-GFMM           & \textbf{2.94} \\ \hline
\end{tabular}
\end{table}
It can be observed that no method is always superior through all datasets, but the average testing error of GFMM using the proposed method is competitive among different methods, such as AGGLO-2 and KNEFMNN. In addition, the IOL-GFMM outperforms other online learning algorithms like Onln-GFMM, FMNN, and EFMNN. To facilitate the conclusion, we rank the performance of approaches, in which the method outputs the lowest error rates on each dataset is ranked first, the second-best method ranks two, and so on. The average ranks through nine datasets of models are shown in Table \ref{avgrank}. It can be easily observed that IOL-GFMM produces the best average performance compared to other types of online learning algorithms of the fuzzy min-max models, as well as being competitive to the batch learning algorithm, i.e., AGGLO-2.

\subsection{A simple ensemble learning approach to tackle the disadvantages of the proposed method}
\begin{figure}[!ht]
\centerline{\includegraphics[width=0.53\linewidth]{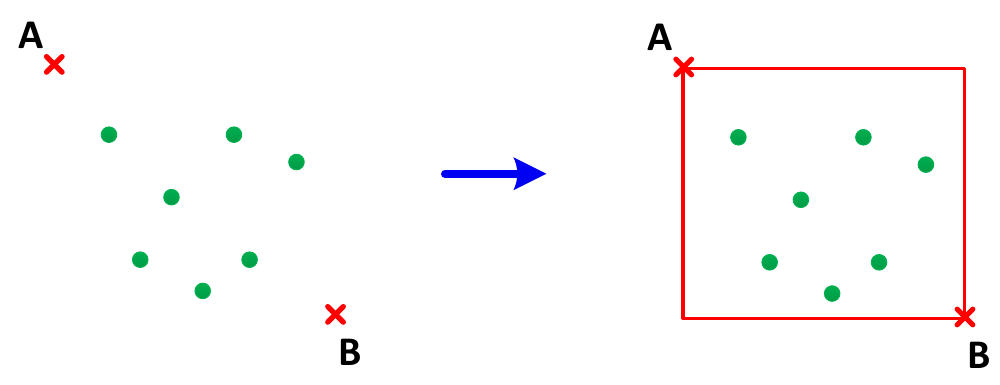}}
\caption{Limitation of the proposed method}
\label{limitation}
\end{figure}

Although the proposed method has shown the effectiveness via experiments, especially in the case of using large values of $\theta$, it is still sensitive to the presentation order of training samples like other online learning algorithms. A disadvantage of the proposed method is shown in Fig. \ref{limitation}. In this case, when samples $A$ and $B$ with the same class label are presented to the GFMM classifier first, they will form a hyperbox. Unfortunately, this hyperbox covers all remaining input patterns of another class. Therefore, when these remaining patterns, which are denoted by dots, come to the network, they do not satisfy the non-overlapping condition between hyperboxes representing different classes. Therefore, all these points form hyperboxes with only one sample. Consequently, all unseen data points with green label located inside the hyperbox with the red label will be incorrectly classified. To overcome this limitation, we should construct small-sized hyperboxes before aggregating these hyperboxes with a larger size, as shown in \cite{Khuat19b}. In addition, if the training data are available, we can use an ensemble method with base learners trained on the same training set but with different data presentation orders in order to overcome the above limitation. By using different orders of training data to build an ensemble model, the limitation presented in Fig. \ref{limitation} is less likely to happen on all base learners, so the ensemble classifier is more robust than the single IOL-GFMM model.

In this experiment, each dataset was split into four folds using the DPS method \cite{Budka13}. For each execution, three folds were used for training the model, and then the performance of the trained model was assessed on the remaining fold. The figures reported in this part are the average values of four testing folds. For each training set, we built eleven GFMM models using the original online learning and IOL-GFMM algorithms, each was trained by randomly shuffling the presentation orders of training data. Next, each model was tested on the remaining testing fold. The average standard deviation for testing errors of models using different learning algorithms on four testing folds is shown in Table \ref{std_shuffle}. Generally, it can be observed that the standard deviation of models trained by the IOL-GFMM is lower than those trained by the Onln-GFMM + Manhattan. The standard deviations for testing error of models using IOL-GFMM are usually lower than 6\% and more stable than those using the original online learning algorithm, especially for \textit{Page blocks} and \textit{Blood transfusion} datasets. However, the IOL-GFMM algorithm is still affected by data presentation orders, especially in the case of using high values of $\theta$. In the case of using small values of $\theta$, online learning algorithms of GFMM are less influenced by data presentation orders.

\begin{table*} [!ht]
\centering
\caption{The Average Standard Deviation (\%) of Algorithms} \label{std_shuffle}
\scriptsize
\begin{tabular}{|c|l|l|c|c|c|c|c|c|c|}
\hline
\textbf{ID} & \textbf{Datasets} & \textbf{Algorithm} & $\theta=0.1$ &  $\theta=0.2$ & $\theta=0.3$ & $\theta=0.4$ & $\theta=0.5$ & $\theta=0.6$ & $\theta=0.7$ \\ \hline
\multirow{2}{*}{1} & \multirow{2}{*}{Blood transfusion} & Onln-GFMM + Manhattan & 3.2647 & 5.4090 & 9.3228 & 10.4414 & 9.8742 & 8.1452 & 8.4970 \\
\cline{3-10}
 &  & IOL-GFMM & 2.7962 & 3.8013 & 4.3108 & 5.1811 & 5.0234 & 4.9881 & 4.5658 \\ \hline
 \multirow{2}{*}{2} & \multirow{2}{*}{Breast Cancer Coimbra} & Onln-GFMM + Manhattan & 0 & 3.2960 & 6.3032 & 5.8891 & 6.4826 & 6.3821 & 6.4206 \\
\cline{3-10}
 &  & IOL-GFMM & 0 & 3.2960 & 5.1123 & 5.020 & 6.0098 & 5.3864 & 5.9864 \\ \hline
 \multirow{2}{*}{3} & \multirow{2}{*}{Haberman} & Onln-GFMM + Manhattan & 2.4569 & 3.8980 & 5.0810 & 5.0356 & 7.1989 & 10.4423 & 12.4998 \\
\cline{3-10}
 &  & IOL-GFMM & 2.4369 & 3.3359 & 3.2321 & 3.5610 & 3.9824 & 4.4686 & 4.3484 \\ \hline
  \multirow{2}{*}{4} & \multirow{2}{*}{Heart} & Onln-GFMM + Manhattan & 0 & 0 & 0 & 0.9722 & 1.9509 & 2.1942 & 1.9381 \\
\cline{3-10}
 &  & IOL-GFMM & 0 & 0 & 0 & 0.4791 & 1.5416 & 1.8697 & 1.620 \\ \hline
 \multirow{2}{*}{5} & \multirow{2}{*}{Page blocks} & Onln-GFMM + Manhattan & 0.3658 & 0.5107 & 2.6319 & 10.6540 & 12.5095 & 15.3907 & 18.3566 \\
\cline{3-10}
 &  & IOL-GFMM & 0.4321 & 0.3701 & 0.4721 &0.4981  & 0.5522 & 0.5154 & 0.5127 \\ \hline
 \multirow{2}{*}{6} & \multirow{2}{*}{Landsat Satellite} & Onln-GFMM + Manhattan & 0.3221 & 0.5147 & 0.9505 & 1.0422 & 1.2635 & 2.3933 & 3.7008 \\
\cline{3-10}
 &  & IOL-GFMM & 0.2922 & 0.4713 & 0.4986 & 0.820 & 0.8033 & 0.7444 & 0.7862 \\ \hline
 \multirow{2}{*}{7} & \multirow{2}{*}{Waveform} & Onln-GFMM + Manhattan & 0 & 0.9443 & 0.8290 & 0.7722 & 0.9021 & 1.4253 & 1.5462 \\
\cline{3-10}
 &  & IOL-GFMM & 0 & 0.9593 & 0.7593 & 0.8792 & 1.1903 & 0.9754 & 1.1182\\ \hline
 \multirow{2}{*}{8} & \multirow{2}{*}{Yeast} & Onln-GFMM + Manhattan & 1.3597 & 1.7120 & 2.5940 & 2.560 &	2.7802 & 3.9750 & 2.6742 \\
\cline{3-10}
 &  & IOL-GFMM & 1.4396 & 1.9833 & 1.9490 & 2.7558 & 2.1458 & 2.5104 & 2.6816 \\ \hline
 \multirow{2}{*}{9} & \multirow{2}{*}{Spherical\_5\_2} & Onln-GFMM + Manhattan & 0.9206 & 0.8130 & 0.7029 & 0 & 0 & 0 & 0\\
\cline{3-10}
 &  & IOL-GFMM & 0.8705 & 0.8130 & 0.7029 & 0 & 0 & 0 & 0 \\ \hline
\end{tabular}
\end{table*}

To reduce the impact of data presentation order, we can build a simple ensemble model of base learners trained on the same training set but with different training sample orders. After that, the predictive results of base learners are aggregated using a majority voting method. Table \ref{avg_ens_shuffle}  shows the average testing errors of eleven base learners and the ensemble model of these eleven base models trained by the IOL-GFMM algorithm. In general, the performance of the ensemble model is better than the single models. With high values of $\theta$, the standard deviation values of base models are usually high, but the ensemble model shows its superior effectiveness in comparison to base estimators. For example, with $\theta = 0.6$, the standard deviations of the base models on \textit{Blood transfusion} and \textit{Breast Cancer Coimbra} datasets are approximately 5\%, whereas the testing errors of ensemble models are lower by about 4\% compared to the average testing errors of the single models. This result confirms that the ensemble method is a suitable approach to overcome the limitations of the proposed method and contribute to building robust learning models.

\begin{table*} [!ht]
\centering
\caption{Average Testing Errors (\%) of IOL-GFMM and the Ensemble Method} \label{avg_ens_shuffle}
\scriptsize
\begin{tabular}{|c|l|l|r|r|r|r|r|r|r|}
\hline
\textbf{ID} & \textbf{Datasets} & \textbf{Algorithm} & $\theta=0.1$ &  $\theta=0.2$ & $\theta=0.3$ & $\theta=0.4$ & $\theta=0.5$ & $\theta=0.6$ & $\theta=0.7$ \\ \hline
\multirow{2}{*}{1} & \multirow{2}{*}{Blood transfusion} & IOL-GFMM & 27.8926 & 27.4307 & 28.3058 & 29.8128 & 29.1687 & 29.0107 & 29.0107 \\
\cline{3-10}
 &  & Ensemble of IOL-GFMMs & 24.3316 & 24.4652 & 24.3316 & 25.1337 & 23.5294 & 24.8663 & 25.4011\\ \hline
 \multirow{2}{*}{2} & \multirow{2}{*}{Breast Cancer Coimbra} & IOL-GFMM & 31.8966 & 28.4483 & 28.6834 & 28.3699 & 29.4671 & 29.3887 & 29.3887 \\
\cline{3-10}
 &  & Ensemble of IOL-GFMMs & 31.8966 & 28.4483 & 25 & 24.1380 & 24.9999 & 25.0000 & 25.8621 \\ \hline
 \multirow{2}{*}{3} & \multirow{2}{*}{Haberman} & IOL-GFMM & 30.0632 & 30.0686 & 29.6282 & 29.5936 & 31.5032 & 31.2057 & 32.6539 \\
\cline{3-10}
 &  & Ensemble of IOL-GFMMs & 30.0666 & 27.1232 & 26.8113 & 25.487 & 28.439 & 27.1318 & 29.4173 \\ \hline
  \multirow{2}{*}{4} & \multirow{2}{*}{Heart} & IOL-GFMM & 21.8503 & 22.2234 & 21.8558 & 21.5914 & 22.1232 & 21.7480 & 20.1932 \\
\cline{3-10}
 &  & Ensemble of IOL-GFMMs & 21.8503	& 22.2234 & 21.8558	& 21.4882 & 21.8503 & 20.3633 & 18.8707 \\ \hline
 \multirow{2}{*}{5} & \multirow{2}{*}{Page blocks} & IOL-GFMM & 4.1244 & 4.0115 & 4.1061 & 4.0762 & 4.3320 & 4.2589 & 4.3570 \\
\cline{3-10}
 &  & Ensemble of IOL-GFMMs & 3.5447 & 3.3986 & 3.3803 & 3.3072 & 3.2523 & 3.5265 & 3.6727 \\ \hline
 \multirow{2}{*}{6} & \multirow{2}{*}{Landsat Satellite} & IOL-GFMM & 10.9006 & 10.5911 & 11.1097 & 12.3019 & 12.9617 & 13.4406 & 13.6680 \\
\cline{3-10}
 &  & Ensemble of IOL-GFMMs & 10.3341 & 9.1064 & 9.5415 & 10.7070 & 11.4218 & 11.4063 & 11.7793 \\ \hline
 \multirow{2}{*}{7} & \multirow{2}{*}{Waveform} & IOL-GFMM &  21.94 & 21.8091 & 18.9382 & 18.9327 & 19.2982 & 19.4727 & 19.4073 \\
\cline{3-10}
 &  & Ensemble of IOL-GFMMs & 21.94 & 18.24 & 15.26 & 15.04 & 15.46 & 15.44 & 15.7 \\ \hline
 \multirow{2}{*}{8} & \multirow{2}{*}{Yeast} & IOL-GFMM & 40.6408 & 43.5354 & 46.2186 & 49.92493 & 53.7171 & 55.7447 & 58.0855 \\
\cline{3-10}
 &  & Ensemble of IOL-GFMMs & 39.0191 & 39.2839 & 41.9788 & 47.1685 & 52.4907 & 54.5810 & 56.0649 \\ \hline
 \multirow{2}{*}{9} & \multirow{2}{*}{Spherical\_5\_2} & IOL-GFMM & 1.2341 & 0.9071 & 1.0502 & 0.7937 & 0.7937 & 0.7937 & 0.7937 \\
\cline{3-10}
 &  & Ensemble of IOL-GFMMs & 1.1967 & 0.3968 & 1.1969 & 0.7937 & 0.7937 & 0.7937 & 0.7937 \\ \hline
\end{tabular}
\end{table*}

\section{Conclusion and future work} \label{conclu}
This paper presented the improved online learning algorithm for the GFMM classifier. The proposed method does not use the contraction step in the training process or reduce the expandable hyperbox candidates for the expansion step. Experimental results indicated that the performance of the proposed approach outperformed the original learning algorithm, especially in the case of using large values of maximum hyperbox size. The new method also showed the robustness to noise in the training sets.

One of the drawbacks of the proposed and original learning algorithms is that they do not handle categorical features effectively because the current membership function is designed for only continuous values. Therefore, one of the intended future research directions will concentrate on addressing this problem in order to build more robust classifiers.


\bibliographystyle{IEEEtran}
\bibliography{myref}

\end{document}